\documentclass{article}
\usepackage{spconf,amsmath,epsfig}
\usepackage[colorlinks,bookmarksopen,bookmarksnumbered,citecolor=red,urlcolor=red,breaklinks]{hyperref}
\usepackage{cleveref}
\usepackage{enumitem}
\usepackage{fancyhdr}
\fancypagestyle{copyright}{%
    \fancyhf{}
    \cfoot{\footnotesize%
        \textcopyright~2024 IEEE. Personal use of this material is permitted. Permission from IEEE must be obtained for all other uses, in any current or future media, including reprinting\slash republishing this material for advertising or promotional purposes, creating new collective works, for resale or redistribution to servers or lists, or reuse of any copyrighted component of this work in other works. This material is referenced by DOI: {\href{}{XXXXXXXXXX}}
    }
}

\title{CLIMATIC \& ANTHROPOGENIC HAZARDS TO THE NASCA WORLD HERITAGE: APPLICATION OF REMOTE SENSING, AI, AND FLOOD MODELLING}

\name{%
Masato Sakai$^1$, Marcus Freitag$^2$, Akihisa Sakurai$^1$, Conrad M Albrecht$^{1,3}$, Hendrik F Hamann$^{1,2}$\thanks{%
\textbf{Author contribution statement}: M.S., M.F., A.S., C.M.A., and H.F.H.\ conceived and planned the study. M.S.\ carried out the data acquisition. M.F.\ and C.M.A.\ worked on the data preparation. M.F.\ carried out the flood modelling. M.S.\ did the field work. M.S., M.F., A.S., C.M.A., H.F.H. jointly wrote the manuscript. M.S.\ and H.F.H.\ supervised the joint study.}
}
\address{%
$^1$ Yamagata University, Yamagata, Japan \\
$^2$ IBM T.J.\ Watson Research Center, Yorktown Heights, NY, USA \\ 
$^3$ German Aerospace Center (DLR), Oberpfaffenhofen, Germany
}

\begin{document}
\maketitle

\begin{abstract}
Preservation of the Nasca geoglyphs at the UNESCO World Heritage Site in Peru is urgent as natural and human impact accelerates. More frequent weather extremes such as flashfloods threaten Nasca geoglyphs. We demonstrate that runoff models based on (sub-)meter scale, LiDAR-derived digital elevation data can highlight AI-detected geoglyphs that are in danger of erosion. We recommend measures of mitigation to protect the famous ``lizard'', ``tree'', and ``hand'' geoglyphs located close by, or even cut by the Pan-American Highway.
\end{abstract}

\begin{keywords}
Archaeology, Nasca Geoglyphs, Artificial Intelligence, Flood Modelling, Remote Sensing, LiDAR, Aerial Imagery, Climatic and Anthropogenic Hazards
\end{keywords}

\thispagestyle{copyright}

\section{Introduction}
\label{sec:intro}

The Nasca Pampa on the southern coast of Peru, South America, is home to the UNESCO World Heritage Site of Lines and Geoglyphs of Nasca. Besides straight-line geoglyphs, trapezoidal and triangular geoglyphs, figurative geoglyphs such as hummingbirds and monkeys approximately 100 meters in size are well known \cite{lumbreras2000contexto}. In recent years, the destruction of geoglyphs has been drawing increased attention \cite{cigna2018tracking,comer2017detecting}. There are two main reasons why geoglyphs are destroyed: human infrastructure and agricultural activities on one hand, and natural flooding on the other hand. For preservation purposes, it is necessary to locate all geoglyphs across the entire Nasca Pampa. However, given the expansive area exceeds 400 km2, conventional surveys would require too much time to identify all endangered geoglyphs. 

Since 2010, we have conducted field surveys of geoglyphs using satellite images, LiDAR data, aerial photos, and drone images. As a result, we have succeeded in identifying more than 1,000 straight lines of geoglyphs distributed in a network \cite{sakai2019centros}. We have also discovered more than 350 new animal and human geoglyphs \cite{NascaGeoglyphs2023}. However, over the past 10 years of field research, we have only been able to survey less than one-third of the entire Nasca Pampa. Given the accelerating pace of human interference with the geoglyphs, it is likely that many will be destroyed even before they are discovered.

In addition to the anthropogenically induced hazards, natural ones such as flooding should not be overlooked. The Nasca Pampa exhibits numerous traces of water runoff, which can be caused by rainwater from the neighboring mountains repeatedly flowing into the Nasca Pampa \cite{NascaLines1990}.
However, there is no precise data on the routes of flooding. It is not known how many geoglyphs have been destroyed by flooding, and which geoglyphs are in danger of being destroyed.

\section{DATA \& METHODS}
\pagestyle{plain}
\label{sec:method}
\begin{figure*}[!t]
\centering
\includegraphics[width=\textwidth]{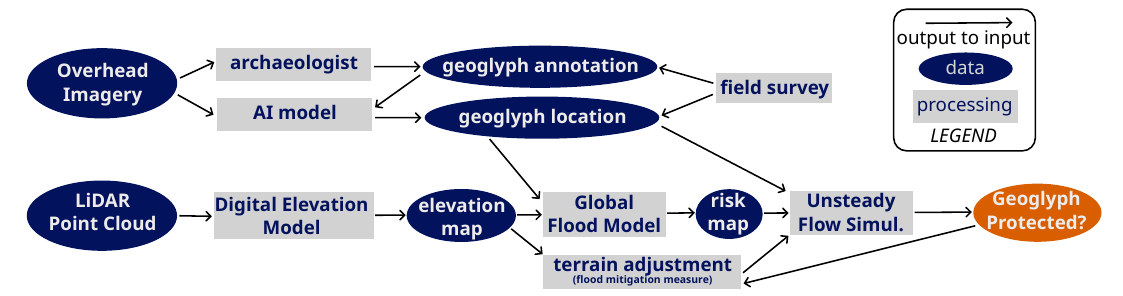}
\caption{Overview of workflow (arrows) to process (boxes) corresponding data (ellipses) in order to protect figurative Nasca geoglyphs where artificial intelligence interacts with remote sensing modalities and domain expert archaeologists.}
\label{fig:workflow}
\end{figure*}

We employ three types of RGB images at different spatial resolutions to locate Nasca geoglyphs: satellite-based imagery (60 cm), airplane-based imagery (10 cm), and drone-based imagery ($<$10 cm). Two types of topographic data have been used: a Digital Elevation Map (DEM, 5 m), and LiDAR-derived DEM (0.5 m).

\Cref{fig:workflow} summarizes our workflow: Airplane-based imagery of the entire Nasca Pampa were obtained. For a subset, geoglyphs are identified and located by archaeologists. After confirmation by a field survey, those manual annotations serve as training data to an Artificial Intelligence (AI) model for identifying geoglyphs in the entire Nasca Pampa. AI-discovered geoglyphs are verified by subsequent field surveys. We identified many geoglyphs invisible from the satellite-based imagery, but resolvable by airplane-based imagery. Continued field research has shown that aerial photographs do not identify all the geoglyphs, nor do they confirm the details of some geoglyphs. Therefore, low-altitude drone photos were taken of the geoglyphs.
The utilization of AI primarily offers these following advantages:
\begin{itemize}[topsep=0pt,itemsep=0ex,partopsep=0ex,parsep=0ex]
    \item significant acceleration of finding geoglyphs directing field surveys to promising areas
    \item consistent approach to survey and identify geoglyphs over extended areas
\end{itemize}
We employed AI for geoglyph detection in the northern part of the Nasca Pampa (approx. 27 km$^2$) where figurative geoglyphs are concentrated. The search for those figures by the naked eye on aerial photographs requires significant human labor. Our experiments demonstrated that with the help of AI models figurative geoglyphs can be found by much shorter. In fact, our proof-of-concept led to the discovery of four unknown geoglyphs. The northern part of the Nasca Pampa is an area subject of intensive field research for Nasca geoglyphs for more than half a century. The discovery of new geoglyphs in such a well surveyed area indicates the effectiveness of our AI-based approach \cite{sakai2023accelerating}.

The impact of flooding on archaeological sites has drawn attention in recent years within the archaeological community \cite{reeder2019preparing,mentzafou2022hydrological}. Our study investigates threats to geoglyphs due to flooding as modelled by elevation from LiDAR data incorporating two runoff simulations. The airborne LiDAR point cloud data of the entire Nasca Pampa were gridded and mosaiced within the IBM GeoDN \slash PAIRS platform \cite{klein2015pairs} as follows:
\begin{enumerate}[topsep=0pt,itemsep=0ex,partopsep=0ex,parsep=0ex]
    \item\label{it:raster} rasterize the LiDAR elevation information at 10 cm resolution (approx. one or zero LiDAR points per pixel),
    \item\label{it:aggregate} aggregate raster layer of \cref{it:raster} to 40 cm adopting the minimum elevation value in 40cm $\times$ 40cm pixel areas, and
    \item fill no-data values in \cref{it:aggregate} by linear interpolation ending up with a Digital Elevation Model (DEM). 
\end{enumerate}
To quantify the flooding danger of known geoglyphs, we are utilizing the DEM \cite{jenson1988extracting,tarboton1991extraction} to:
\begin{enumerate}[topsep=0pt,itemsep=0ex,partopsep=0ex,parsep=0ex]
    \item computationally fill in depressions (sinks) that may exist due to lakes, sink-holes, or karst geology,
    \item calculate ``flow direction'' and ``steepness'' for every pixel by finding the lowest-elevation neighbor,
    \item determine the watershed boundary
    \item compute the layer of ``flow accumulation'' summarizing the area above each pixel that drains through that pixel,
    \item vectorize the ``flow accumulation'' by using ``flow direction'' to link pixels together and to create nodes where links combine.
\end{enumerate}
Since flow accumulation can extend far into the Andes Mountains, for which we did not acquire LIDAR data, we employed a lower-resolution (5 m) DEM to connect the vectorized LIDAR flow accumulation with a similar flow accumulation from the Andes mountains. The resulting vectorized network is concentrated in the flow channel bottoms.

To simulate desert flash floods where water levels rise by, e.g., 10 cm, we employ a rolling 2D Gaussian kernel of 41$\times$41 pixels to distribute the “flow accumulation” attribute over the neighboring 41$\times$41 pixels as long as their elevation is less than 10cm above the center pixel. A continuous ``flooding flow accumulation'' layer is created by a ``maximum'' aggregation for the above rolling window operation. For endangered geoglyphs identified by this flood model, we simulated a time-resolved 2D unsteady flow using HEC-RAS \cite{alzahrani2017application}.

\section{RESULTS \& DISCUSSION}
\label{sec:results}
\begin{figure*}[!t]
\centering
\includegraphics[width=\textwidth]{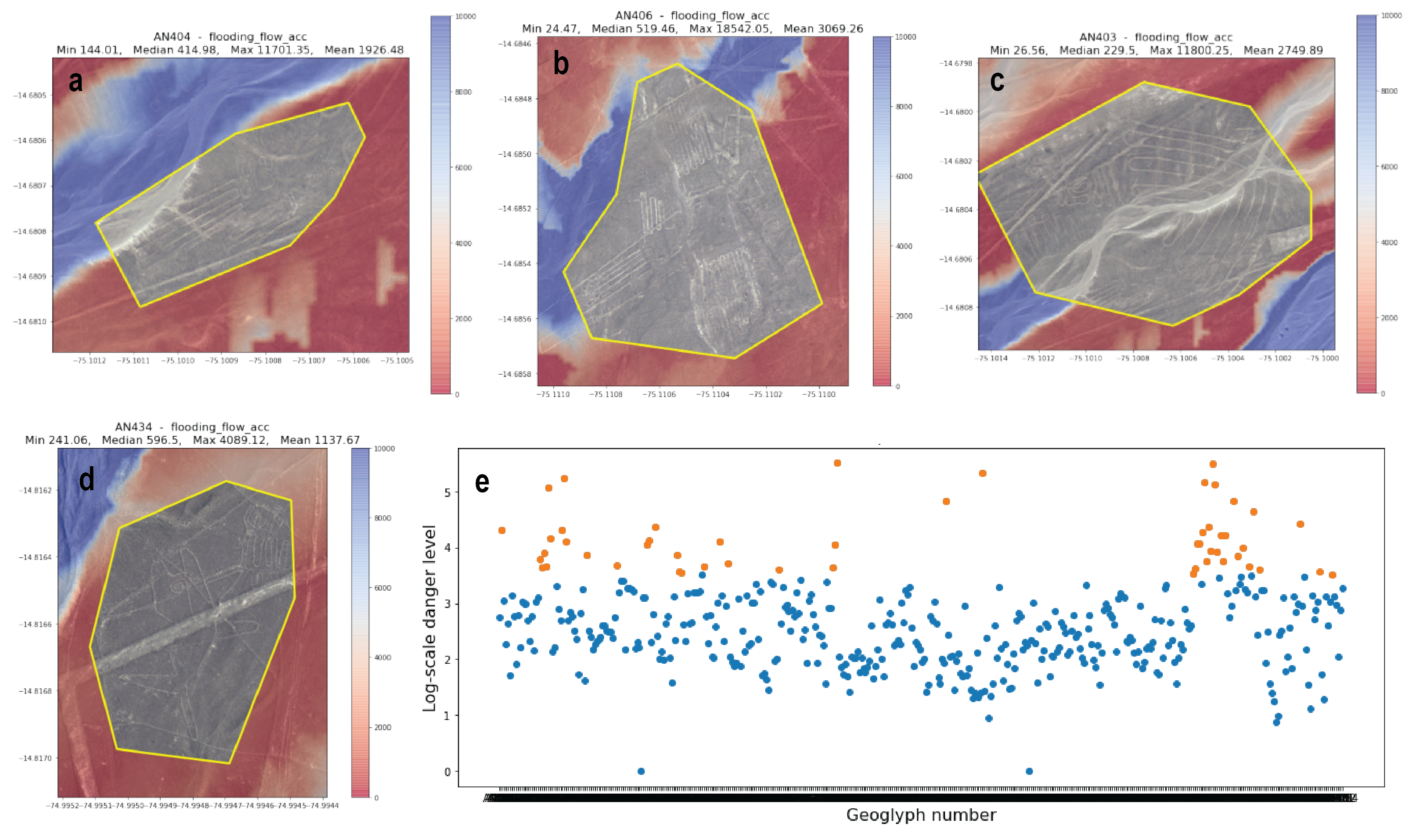}
\caption{Modelled erosion danger due to flooding. \textbf{(a-d)} Examples of the flooding flow accumulation as a proxy for the danger of erosion. The color-scale indicates potentially ``safe'' levels of flow accumulation are below 3257, which corresponds to a flow-accumulation area of 100m $\times$ 100m (red). Potentially ``unsafe'' levels of flow accumulation are shown in blue\slash white. \textbf{(e)} Maximum pixel-level danger level for each geoglyph on a log-scale. Eroding geoglyphs with pixels above the ``safe'' level are shown in orange.}
\label{fig:flooding}
\end{figure*}
\begin{figure*}[!t]
\centering
\includegraphics[width=\textwidth]{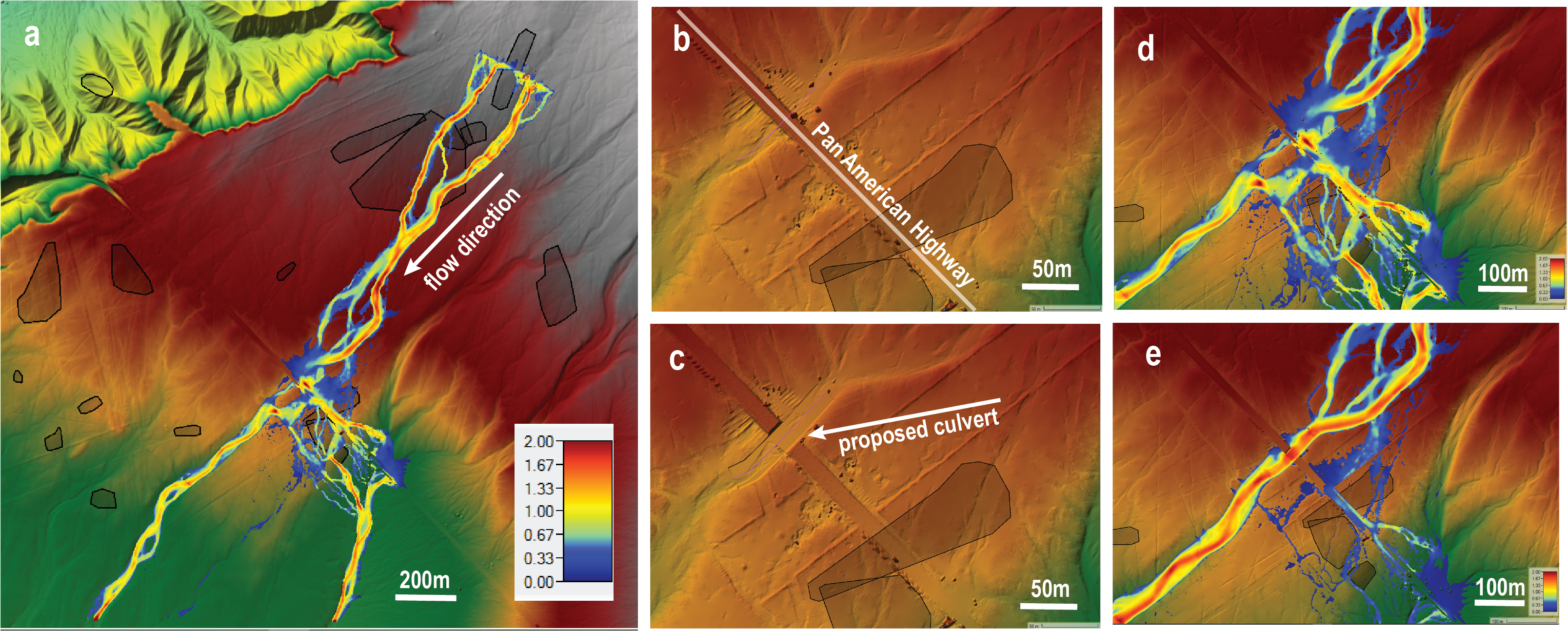}
\caption{2D unsteady flow simulation using HEC-RAS. \textbf{(a)} Flow velocity in m/s for the entire field of view of the simulation. The LIDAR-derived elevation map is shown in false-color. Location with known geoglyphs are shaded. Inflow boundary condition is 20 cubic meters rain per second applied along a line in the top-right corner. \textbf{(b)} Topography around the Pan-American highway cutting from the top-left corner to the bottom-right. \textbf{(c)} Topography altered to simulate an added water channel (culvert) under the highway. \textbf{(d)} Unmitigated flow intersecting three famous geoglyphs (lizard, tree, and hand). \textbf{(e)} Mitigated flow mostly takes the path through the culvert under the highway.}
\label{fig:hecras}
\end{figure*}

A survey combining remote sensing, AI, and archaeology led to the discovery of a large number of new geoglyphs, dramatically improving our understanding of the geospatial distribution of geoglyphs in the Nasca pampa. A paper detailing these results is under submission. Here we focus on assessing the risk of those geoglyphs eroding due to flooding. 

The erosion danger of each geoglyph is modelled using the ``flooding flow accumulation'' as described in the methods section. \Cref{fig:flooding} shows four examples of this parameter around four large geoglyphs. In some of the geoglyphs, such as \cref{fig:flooding}a,c, where we find parts of the geoglyph above the threshold, we can discern tracks of past flash floods that confirm our model. In \cref{fig:flooding}b parts of the geoglyph are eroded in high-danger areas of the geoglyph (wing). The ``whale'' geoglyph \cref{fig:flooding}d can be considered entirely safe from flash-floods not exceeding 10cm in height due to its location on higher ground---the line crossing the whale is a human-made Nasca Line. \Cref{fig:flooding}e summarizes the known geoglyphs according to their maximum flooding flow accumulation. The ones highlighted in orange warrant further scrutiny in order to protect them from further erosion. 

One limitation of the model above is that it cannot deal with situations where the water flow splits into multiple streams during flooding. To properly simulate time-resolved 2D unsteady flow, we apply HEC-RAS, \cite{alzahrani2017application}. It is not as automated as our model, but for the limited geoglyphs we identified as potentially threatened, it can deliver additional insights. 

\Cref{fig:hecras} shows the Pan-American Highway, built in 1938 before the figurative geoglyphs were discovered, cutting through one of the geoglyphs (the lizard) and affecting rain-water runoff during flash-floods. It is apparent that two additional geoglyphs (the tree and hand geoglyphs) are now in the path of the water diverted by the highway. One of them (the hand geoglyph) was already affected by a strong rain event in 2019. To prevent further destruction, we propose restoring the original path of the water by building a culvert under the highway. 

To examine the simulation results, we conducted a field survey with specialists from the Peruvian Ministry of Culture. As a result, the field survey confirmed that these figurative geoglyphs were very likely to be destroyed by water currents, as per the simulation.

\section{Conclusions \& Perspectives}

To summarize, we were able to identify figurative geoglyphs that are in danger of destruction due to heavy rains. We were also able to ascertain which geoglyphs are most likely to be impacted. Our research has revealed the details of how the rain that fell on the mountains near the Nasca Pampa flowed into the Pampa and eroded parts of the geoglyphs. It is particularly noteworthy that some of the most-affected geoglyphs are distributed along the highway. It is clear that these geoglyphs were damaged not only by natural disasters, but also by anthropogenic influences. We proposed a possible mitigation measure to prevent further damage of the geoglyphs near the Pan-American Highway. A comprehensive plan will be developed to protect the geoglyphs that are most likely to be destroyed. Activities to protect them will be carried out in collaboration with the Peruvian Ministry of Culture.

\small
\section{ACKNOWLEDGEMENT}
\label{sec:ref}

This work was supported by Japan Society for the Promotion of Science KAKENHI (grant no. 20H00041), Yamagata University YU-COE(S)(S-4), and the Helmholtz Association through the Framework of \textit{HelmholtzAI}, grant ID: \texttt{ZT-I-PF-5-01} -- \textit{Local Unit Munich Unit @Aeronautics, Space and Transport (MASTr)}. This work was also generously supported by IBM Research through a Joint Study agreement.

\bibliographystyle{IEEEbib}
\bibliography{refs}

\begin{thebibliography}{10}

\bibitem{lumbreras2000contexto}
L.~G. Lumbreras,
\newblock {\em {Formulación de los lineamientos para la elaboración de un
  Plan de manejo de las líneas de Nasca 1: contexto arqueológico}},
\newblock {Instituto Nacional de Cultura del Perú \& UNESCO, Lima}, 2000.

\bibitem{cigna2018tracking}
F. Cigna and D. Tapete,
\newblock ``Tracking human-induced landscape disturbance at the nasca lines
  unesco world heritage site in peru with cosmo-skymed insar,''
\newblock {\em Remote Sensing}, vol. 10, no. 4, pp. 572, 2018.

\bibitem{comer2017detecting}
D.~C. Comer, B.~D. Chapman, and J.~A. Comer,
\newblock ``Detecting landscape disturbance at the nasca lines using sar data
  collected from airborne and satellite platforms,''
\newblock {\em Geosciences}, vol. 7, no. 4, pp. 106, 2017.

\bibitem{sakai2019centros}
M. Sakai, J. Olano, and H. Takahashi,
\newblock {\em Centros de L{\'\i}neas y Cer{\'a}mica en las Pampas de Nasca,
  Per{\'u}, hasta el a{\~n}o 2018},
\newblock Yamagata University Press, Yamagata, 2019.

\bibitem{NascaGeoglyphs2023}
M. Sakai,
\newblock {\em Nasca geoglyphs: spatial distribution rule, making purpose, and
  protection (in Japanese)},
\newblock Embassy of the Republic of Peru in Japan, Tokyo, 2023.

\bibitem{NascaLines1990}
{Aveni, A. F. (ed.)},
\newblock {\em The lines of Nazca},
\newblock American Philosophical Society, Philadelphia, 1990.

\bibitem{sakai2023accelerating}
M. Sakai, Y. Lai, J.~O. Canales, M. Hayashi, and K. Nomura,
\newblock ``Accelerating the discovery of new nasca geoglyphs using deep
  learning,''
\newblock {\em Journal of Archaeological Science}, vol. 155, pp. 105777, 2023.

\bibitem{reeder2019preparing}
L.~A. Reeder-Myers and M.~D. McCoy,
\newblock ``Preparing for the future impacts of megastorms on archaeological
  sites: An evaluation of flooding from hurricane harvey, houston, texas,''
\newblock {\em American Antiquity}, vol. 84, no. 2, pp. 292--301, 2019.

\bibitem{mentzafou2022hydrological}
A. Mentzafou and E. Dimitriou,
\newblock ``Hydrological modeling for flood adaptation under climate change:
  the case of the ancient messene archaeological site in greece,''
\newblock {\em Hydrology}, vol. 9, no. 2, pp. 19, 2022.

\bibitem{klein2015pairs}
L.~J. Klein, F.~J. Marianno, C.~M. Albrecht, M. Freitag, S. Lu, N. Hinds, X.
  Shao, S.~B. Rodriguez, and H.~F. Hamann,
\newblock ``Pairs: A scalable geo-spatial data analytics platform,''
\newblock in {\em 2015 IEEE International Conference on Big Data (Big Data)}.
  IEEE, 2015, pp. 1290--1298.

\bibitem{jenson1988extracting}
S.~K. Jenson and J.~O. Domingue,
\newblock ``Extracting topographic structure from digital elevation data for
  geographic information system analysis,''
\newblock {\em Photogrammetric engineering and remote sensing}, vol. 54, no.
  11, pp. 1593--1600, 1988.

\bibitem{tarboton1991extraction}
D.~G. Tarboton, R.~L. Bras, and I. Rodriguez-Iturbe,
\newblock ``On the extraction of channel networks from digital elevation
  data,''
\newblock {\em Hydrological processes}, vol. 5, no. 1, pp. 81--100, 1991.

\bibitem{alzahrani2017application}
A.~S. Alzahrani,
\newblock ``Application of two-dimensional hydraulic modeling in riverine
  systems using hec-ras,''
\newblock M.S. thesis, University of Dayton, 2017.

\end{thebibliography}

\end{document}